\documentclass[]{ecai}  

\usepackage{graphicx}
\usepackage{latexsym}

\paperid{682}        

\usepackage{microtype}
\usepackage{soul}
\usepackage[utf8]{inputenc}
\usepackage{subcaption}
\usepackage{xcolor}
\usepackage{amsmath}
\usepackage{amsthm}
\usepackage{amssymb}
\usepackage{booktabs}
\usepackage{csquotes} \MakeOuterQuote{"}
\usepackage[ruled,vlined]{algorithm2e}
\usepackage[hidelinks]{hyperref}
\usepackage[nameinlink,capitalise]{cleveref}
\creflabelformat{equation}{#2#1#3}  
\usepackage[nocompress]{cite}
\graphicspath{{fig/}}
\usepackage[normalem]{ulem}
\newtheorem{theorem1}{Theorem}
\newtheorem{definition}{Definition}
\newtheorem{lemma}{Lemma}
\newcommand{\AlgComment}[1]{ \hfill {\small$\triangleright$\ {#1}}}

\begin{document}

\begin{frontmatter}

\title{Diverse, Top-k, and Top-Quality Planning Over Simulators}

\author[A,B]{\fnms{Lyndon}~\snm{Benke}}
\author[B]{\fnms{Tim}~\snm{Miller}}
\author[A]{\fnms{Michael}~\snm{Papasimeon}}
\author[B]{\fnms{Nir}~\snm{Lipovetzky}}

\address[A]{Defence Science and Technology Group, Australia}
\address[B]{The University of Melbourne, Australia}

\begin{abstract}
    Diverse, top-$k$, and top-quality planning are concerned with the generation of sets of solutions to sequential decision problems. Previously this area has been the domain of classical planners that require a symbolic model of the problem instance. This paper proposes a novel alternative approach that uses Monte Carlo Tree Search (MCTS), enabling application to problems for which only a black-box simulation model is available. We present a procedure for extracting bounded sets of plans from pre-generated search trees in best-first order, and a metric for evaluating the relative quality of paths through a search tree. We demonstrate this approach on a path-planning problem with hidden information, and suggest adaptations to the MCTS algorithm to increase the diversity of generated plans. Our results show that our method can generate diverse and high-quality plan sets in domains where classical planners are not applicable.
\end{abstract}

\end{frontmatter}

\section{Introduction}

Automated planning is a model-based approach to the discovery of solutions to sequential decision problems. In typical applications, a search algorithm generates a single strategy or action sequence to solve a specific planning task.
This paper explores the generation of bounded \emph{sets} of plans, described as \emph{diverse}, \emph{top-$k$}, or \emph{top-quality} planning when the set is bounded by diversity, cardinality, or quality, respectively.
Planners that produce plan sets have applications in a wide range of fields, including planning for problems with incomplete or unknown user preferences \cite{nguyen_generating_2012}, scenario prediction for risk management \cite{sohrabi_ai_2018}, plan recognition \cite{sohrabi_state_2017}, plan repair \cite{fox_plan_2006}, and explanation generation \cite{boddy_course_2005}.
Our research is motivated by the discovery of novel agent behaviours in the context of operations research \cite{ramirez_integrated_2018,masek_discovering_2018,papasimeon_multi-agent_2021}, recently proposed as an input to deceptive mission planning \cite{benke_modelling_2021}.

Previous approaches to generating plan sets have focused on classical planning, typically employing repeated calls to off-the-shelf solvers \cite{srivastava_domain_2007,riabov_new_2014,katz_novel_2018,speck_symbolic_2020}.
Such approaches take advantage of structural features of the planning problem for efficient solution discovery, and hence require a symbolic model of the domain and problem instance in a declarative planning language such as STRIPS or PDDL.
We consider problems for which a declarative model is \emph{not} available, and classical planners cannot be used off-the-shelf as action effects are not known before execution. Instead, a black-box simulator provides a sample of a successor state and reward given an input state and action.
Such problems are particularly widespread in operations research, where Monte Carlo simulation is used to study complex sociotechnical systems \cite{papasimeon_multiagent_2018}. These simulations include high-fidelity models of cyber-physical systems, such as fighter aircraft, that are impractical or impossible to model in symbolic planning languages.

Monte Carlo Tree Search (MCTS) is an alternative approach to planning that has achieved considerable success in complex domains with large branching factors \cite{browne_survey_2012}.
Unlike classic planners, MCTS does not require a symbolic model of the planning problem, and can be used when only a generative model is available. 
While it is commonly used for online, adversarial planning, building a new search tree at each time-step to estimate the best single action from a given state, it has also proven to be effective in offline and single-player contexts, such as those studied in previous research on diverse, top-$k$ and top-quality planning problems.

In this paper, we present a novel method for extracting bounded sets of plans from pre-generated search trees.
While we focus on MCTS, our approach may be applied to any tree search algorithm that backpropagates node values.
We additionally propose a metric for the relative quality of paths in a search tree, based on expected return, that guarantees best-first ordering.
We present proofs for soundness and completeness, and discuss modifications to the MCTS algorithm to increase the diversity of generated plans.
Finally, we provide an empirical evaluation in a path-planning scenario with hidden information, and discuss the practicality of our approach for diverse, top-$k$ and top-quality planning.
The experimental results suggest that diverse planning is most effective for problems with moderate to high levels of hidden information, where any decrease in efficiency is offset by an increased probability of success.

In line with related work, this paper considers the generation of plan sets in the context of deterministic single-player planning problems. The extension of our approach to multi-agent problems is left for future work.

\section{Related Work} \label{sec:related-work}

Previous works have explored top-$k$, top-quality and diverse planning using either generalisations of A* search known as K* \cite{riabov_new_2014}, or replanning approaches that repeatedly reformulate the planning task to forbid a growing set of solutions \cite{katz_novel_2018,speck_symbolic_2020}.
Approaches based on K* require a heuristic function that may be difficult to obtain, and are outperformed by replanning approaches for small values of $k$, while replanning approaches incur significant additional cost as the size of the plan set increases.
Additionally, replanning approaches require a symbolic model of the domain and problem instance in a declarative planning language such as STRIPS or PDDL, and as a result cannot be applied to problems for which only a black-box simulation model is available. This limitation also applies when deriving domain-independent heuristics for K*-based approaches.

Recent works have proposed addressing diversity separately to top-$k$ and top-quality planning, by post-processing an existing set of plans.
One approach proposes a greedy method, iteratively selecting plans that maximise the diversity of the growing set \cite{katz_reshaping_2020}, while another finds an optimally diverse subset by solving a mixed-integer linear programming problem \cite{katz_bounding_2022}.
However, since these approaches operate on pre-generated sets of plans, they cannot be applied over simulators without first using a method such as the one presented in this paper to generate a top-$k$ or top-quality set of plans.

MCTS has been applied to single-player problems in a variety of domains, in some cases outperforming classical methods and achieving a level of performance comparable to the best hand-crafted approaches \cite{schadd_single-player_2008,bjornsson_cadiaplayer_2009}.
Relevant applications have included constructing an MCTS tree offline and querying the subsequent tree during gameplay \cite{auger_multiple_2011}, and employing offline MCTS to generate a Pareto set of possible winning positions \cite{perez_online_2013}.
While the extraction of a single optimal plan from an MCTS tree has been considered in the context of nested Monte Carlo search \cite{cazenave_high-diversity_2016} and hierarchical task decomposition \cite{parascandolo_divide-and-conquer_2020}, to our knowledge no previous work has explored the generation of bounded \emph{sets} of plans using MCTS; our synthesis of these concepts appears to be unique.

\section{Background and Preliminaries}

\subsection{Markov Decision Processes}
A planning problem that provides rewards rather than explicit goals is typically formulated as a Markov Decision Process (MDP). In an MDP, an agent interacts with an environment by taking actions and receiving rewards. The objective of the agent is to maximise the expected cumulative reward it receives over time.
An MDP is defined by a tuple $\langle S, A, T, R \rangle$, where $S$ is a set of possible states, $A$ is a set of actions available to the decision-maker, $T(s, a, s')$ is a transition model, and $R(s)$ is a reward function. The transition model determines the probability of reaching state $s'$ after taking action $a$ in state $s$, while the reward function defines the reward received after a transition. For deterministic problems, such as those considered here, taking action $a \in A$ in state $s \in S$ always results in a unique next state $s' \in S$.
A solution to an MDP is a \emph{plan} or \emph{policy} that maximises the expected reward for the agent. 
A plan is a sequence of actions $\pi = \{ a_0, \dots, a_n \}$ that generates one or more sequences of states $\{ s_0, \dots, s_n \}$, where $s_0$ is the initial problem state.
A policy is a mapping from states to actions that specifies the action to select from any given state $s \in S$.

\subsection{Search Trees}
A \emph{game tree} is a hierarchical graph structure that is used to model and analyse sequential decision problems such as MDPs, where each node represents a game state, and each edge a possible transition from one state to another.
A \emph{complete} game tree includes every node that is reachable from the initial state, capturing the full search space for the decision problem. Due to computational limitations, in practical applications a partial game tree or \emph{search tree} is usually used instead, representing a subgraph of the complete tree that is constructed through repeated sampling of the search space.
For example, A* search builds a partial tree by selecting nodes based on a heuristic, while Monte Carlo Tree Search employs simulated playouts.
A plan in a search tree corresponds to a path from the tree root to a leaf node.

\subsection{Monte Carlo Tree Search (MCTS)} \label{sec:mcts-overview}
MCTS is a family of algorithms that apply stochastic game-tree search to sequential decision problems. A search tree is generated that contains nodes representing the visited subset of \textit{state-action} pairs, with each node maintaining an approximation of the value of a particular transition.
During each iteration, the MCTS algorithm selects a leaf node by descending the tree from the root according to a \emph{tree policy}, expands the selected node if possible by adding a new child node, executes a simulated playout according to a \emph{default policy}, and backpropagates the result by updating the statistics for each of the visited nodes.
After a predetermined time or number of iterations, the action corresponding to the best child of the root node is returned.

The tree policy balances exploitation of nodes with high value estimates against exploration of nodes with low visit counts by recursively applying a multi-armed bandit policy, such as UCB1 \cite{auer_finite-time_2002} which selects the node that maximises the value:
    \begin{equation} \label{eq:ucb1}
        \text{UCB1}(\sigma) = Q(\sigma) + C \sqrt{\frac{2 \ln n}{n_\sigma}}
    \end{equation}
where $Q(\sigma)$ is the value estimate for node $\sigma$, $n_\sigma$ and $n$ are the number of times the node and its parent have been visited respectively, and $C$ is a positive constant balancing exploration and exploitation.
The node value considered by the tree policy, $Q(\sigma)$, is the average reward based on repeated Monte Carlo playouts, representing an estimate of the expected return for choosing the corresponding action in the previous state:
    \begin{equation} \label{eq:node-value}
        Q(\sigma) = Q(s, a) = \frac{z_\sigma}{n_\sigma}
    \end{equation}
where $s$ and $a$ are the state and action leading to the node, and $z_\sigma$ is the total reward received at $\sigma$.
As results are backpropagated, the $q$-value for each node also represents the weighted average of the values of its set of child nodes $C(\sigma)$:
    \begin{equation} \label{eq:node-value-average}
        Q(\sigma) = \frac{1}{\sum_{\sigma' \in C(\sigma)}n_{\sigma'}} \sum_{\sigma' \in C(\sigma)} n_{\sigma'} Q(\sigma')
    \end{equation}

Previous research has demonstrated that MCTS performance can be improved in single-player games by choosing the final move according to the \emph{maximum} playout result rather than the average, since there is no adversary to prevent optimistic play \cite{bjornsson_cadiaplayer_2009}.
In such cases, the $q$-value that is used for action selection is the maximum child value:
    \begin{equation} \label{eq:node-value-deterministic}
        Q_{max}(\sigma) = \max_{\sigma' \in C(\sigma)}Q(\sigma')
    \end{equation}

For games that award a win or loss to the player, $Q(\sigma)$ may be considered an estimate of the probability that the player will win after visiting the node, when following the tree and default policies.

\subsection{Diverse, Top-k and Top-Quality Planning} \label{sec:planning-defs}
The generation of multiple solutions to a planning problem, rather than a single optimal plan, is known as \emph{diverse}, \emph{top-$k$}, or \emph{top-quality} planning, depending on the attributes used to bound the plan set. Where top-$k$ planning restricts the set of plans to a fixed size, top-quality planning enforces a minimum solution quality, and diverse planning adds a constraint on plan similarity.
We note here the distinction between a \emph{plan} in the context of classical planning, being a sequence that successfully achieves an explicit goal, and a \emph{plan} in the context of an MDP, being any sequence with an associated expected return.

Top-$k$ planning addresses the problem of finding a finite set of plans of size $k$, such that no plan with greater quality exists outside of the set according to some measure of plan quality provided by the user \cite{deng_efficient_2013,riabov_new_2014,speck_symbolic_2020}. Top-$k$ planning extends the well-studied \emph{$k$ shortest paths} problem, and is typically applied to problems requiring sets of high-quality but similar plans, such as plan repair \cite{fox_plan_2006}.
Single planning may be considered a special case of top-$k$ planning, for which $k=1$.
Drawing on classical planning \cite{speck_symbolic_2020}, a top-$k$ planning problem may be defined for an MDP as follows:
    \begin{definition}[Top-$k$ planning problem] \label{def:top-k-planning} \ \\
        Given a planning problem $\Pi = \langle S, A, T, R \rangle$, measure of plan quality $Q_{plan}(\pi)$, and natural number $k$, find a subset $\mathcal{P}$ in the set of all plans $\mathcal{P}_\Pi$ such that:
        \begin{enumerate}
            \item there exists no plan $\psi \in \mathcal{P}_\Pi$ with $\psi \notin \mathcal{P}$ that has quality $Q_{plan}(\psi)$ greater than some plan $\pi \in \mathcal{P}$, and
            \item $|\mathcal{P}| = k$ if $|\mathcal{P}_\Pi| \geq k$, or $|\mathcal{P}| = |\mathcal{P}_\Pi|$ otherwise.
        \end{enumerate}
    \end{definition}
Note that $k$ is regarded as a constraint rather than a requirement \cite{nguyen_generating_2012}, where some works consider the problem to be unsolvable if fewer than $k$ plans are found.

Top-quality planning is a variation on top-$k$ planning wherein the plan set is subject to limits on quality, rather than cardinality \cite{roberts_evaluating_2014,vadlamudi_combinatorial_2016,katz_top-quality_2020}.
This approach is useful when the user does not require a specific number of plans, but has well-defined constraints that bound acceptable solutions.
Drawing on classical planning \cite{katz_top-quality_2020}, a top-quality planning problem may be defined for an MDP as follows:
    \begin{definition}[Top-quality planning problem] \label{def:top-quality-planning} \ \\
        Given a planning problem $\Pi = \langle S, A, T, R \rangle$, measure of plan quality $Q_{plan}(\pi)$, and quality constraint $q$, find the set of plans $\mathcal{P} \subset \mathcal{P}_\Pi$, where $\mathcal{P} = \lbrace \pi \in \mathcal{P}_\Pi \; | \; Q_{plan}(\pi) \geq q \rbrace$.
    \end{definition}

Diverse planning considers the problem of generating sets of plans that are meaningfully different according to some measure of plan distance. Diversity is defined as the minimum or average pairwise distance within a set of plans, and has been approached using both qualitative measures \cite{myers_generating_1999}, and domain-independent, quantitative measures \cite{nguyen_generating_2012}. Diverse planning is typically combined with a bound on plan set cardinality \cite{nguyen_generating_2012,bryce_landmark-based_2014}, in some cases with an additional bound on plan quality \cite{roberts_evaluating_2014,katz_reshaping_2020}.
We consider the problem of maximising quality given a bound on diversity, in contrast to recent works that maximise diversity given a bound on quality \cite{katz_reshaping_2020,katz_bounding_2022}. This approach is more applicable to our domains of interest, which typically require the highest-quality plans with sufficient diversity, rather than the most diverse set of plans.
A quality-optimal, diversity-bounded top-$k$ planning problem may be defined for an MDP as follows:
    \begin{definition}[Diverse top-$k$-quality planning problem] \label{def:diverse-planning} \ \\
        Given a planning problem $\Pi = \langle S, A, T, R \rangle$, natural number $k$, measure of plan quality $Q_{plan}(\pi)$, quality constraint $q$, measure of minimum pairwise distance $D(\pi, \mathcal{P})$, and distance constraint $d$, find a \emph{maximal} subset $\mathcal{P}$ in the set of all plans $\mathcal{P}_\Pi$ such that:
        \begin{enumerate}
            \item there exists no plan $\psi \in \mathcal{P}_\Pi$, $\psi \notin \mathcal{P}$ that has diversity $D(\psi, \mathcal{P}) \geq d$ and quality $Q_{plan}(\psi)$ greater than some plan $\pi \in \mathcal{P}$, and
            \item for all $\pi \in \mathcal{P}$, $D(\pi, \mathcal{P} - \pi) \geq d$, and
            \item for all $\pi \in \mathcal{P}$, $Q_{plan}(\pi) \geq q$, and
            \item $|\mathcal{P}| \leq k$.            
        \end{enumerate}
    \end{definition}
The solution to a diverse planning problem is hence a Pareto set balancing diversity and quality.
Diverse planning by this definition may be considered a generalisation of top-$k$ and top-quality planning, since either alternative may be achieved by setting the diversity constraint to zero.

\section{Plan Extraction from Search Trees} \label{sec:plan-extraction}

MCTS is typically used for online decision-making, informing the selection of a \emph{single action} by generating a new partial tree at each time step that provides an estimate of the value of each transition from the current state.
However, MCTS may also be used to generate \emph{plans} by extracting action sequences from the search tree, where a plan in this context corresponds to any path from the tree root to a leaf node.
In the simplest case, the optimal plan may be extracted by starting at the root node and recursively selecting the child node with the maximum $q$-value until a leaf node is reached, returning the corresponding action sequence (plan $\alpha$ in \cref{fig:plan-quality-tree}).
Generating a \emph{set} of plans is more complex, requiring a method for extracting and comparing alternative plans in a search tree without the benefit of explicit plan costs. The following sections introduce a measure for the relative quality of paths through a search tree, and a process for extracting plans in best-first order subject to various constraints.

\subsection{Plan Quality Metric} \label{sec:plan-quality}
\begin{figure}[t]
    \centering
    \includegraphics[width=1.0\linewidth]{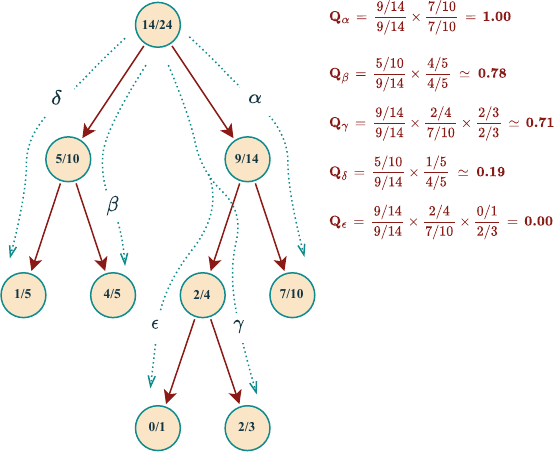}
    \caption{Illustration of relative plan quality calculation for a search tree with backpropagated rewards. Plan quality is defined by the reduction in expected return attributed to each node on the path (\cref{eq:plan-quality}).}
    \label{fig:plan-quality-tree}
\end{figure}

Unlike in classical planning, where the quality of a plan is naturally defined by the sum of action costs \cite{katz_top-quality_2020}, in standard MCTS implementations rewards are assumed to be received only in terminal states. Nodes instead track the backpropagated results of simulated playouts, typically in the form of a win or loss.
We propose a quality measure that is based on expected return relative to the optimal plan, rather than plan cost:
\begin{definition}[Relative Plan Quality] \label{def:plan-quality} \ \\
    Given a node sequence (plan) representing a path through a search tree, $\pi = \{\sigma_0, \dots, \sigma_n\}$, the relative quality of $\pi$ is defined as the product of the regret due to suboptimal action choices, quantified as the reduction in expected return attributed to each node in the plan:
    \begin{equation} \label{eq:plan-quality}
        Q_{plan}(\pi) = \prod_{i = 0}^{n-1} \frac{Q(\sigma_{i+1})}{\max_{\sigma' \in C(\sigma_i)}Q(\sigma')}
    \end{equation}
    where $Q(\sigma) \in \mathbb{R^+}$ is the backpropagated quality of node $\sigma$, and $C(\sigma_i)$ is the set of child nodes of $\sigma_i$.
\end{definition}
An example of plan quality calculation for a search tree is provided in \cref{fig:plan-quality-tree}.
It follows that the top plan in any tree has a relative quality equal to one, since by definition the optimal plan selects the best available action in each state.
This approach provides a natural basis for comparing plans with different lengths: given two plans of unequal length $p$ and $q$ with $|p| < |q|$, the quality of $p$ is equal to the quality of a plan $p'$ padded with optimal actions such that $|p'| = |q|$.

The \emph{absolute} quality for any plan, corresponding to the expected return at the initial state when following the tree policy, may be calculated by multiplying the relative plan quality by the $q$-value of the root node. This absolute quality may vary greatly, since it depends on both the problem definition and the growth of the MCTS tree. By defining plan quality \emph{relative} to the optimal plan, the metric defined above can be used to bound plan sets without knowledge of the actual $q$-value estimates in the tree. 

Since the relative quality of a plan is defined as the product of the quality of its action choices, it follows that the quality for any sub-plan must be greater than or equal to the quality of the complete plan:
    \begin{lemma} \label{lem:subsequence-quality}
        If $\pi$ is a sequence of nodes $\{\sigma_0, \dots, \sigma_n\}$ representing a path through a search tree with backpropagated node values, then any subsequence $\psi \subset \pi$ has quality $Q_{plan}(\psi) \geq Q_{plan}(\pi)$.
    \end{lemma}
    \begin{proof}
        Let $Q_{plan}(\pi)$ be the quality of $\pi$ as defined by \cref{eq:plan-quality}. Since $Q_{plan}(\pi)$ is the product of the $q$-value ratio corresponding to each node $\sigma \in \pi$, and each ratio lies in the range [0, 1] as the denominator cannot be greater than the divisor, 
        then $\forall(\psi \subset \pi): Q_{plan}(\psi) \geq Q_{plan}(\pi)$.
    \end{proof}

\subsection{Plan Extraction Algorithm}
We present a process for extracting top plans from a search tree in \cref{algo:plan-extraction}.
The algorithm is general and can be applied to diverse, top-$k$, or top-quality planning problems by adjusting the bounding constraints and metrics accordingly.
The search tree is traversed by incrementally expanding plans in order of quality, representing a best-first search of the plan space defined by the pre-generated tree.
During execution of the algorithm, the open list $\mathcal{O}$ contains a queue of node sequences ordered by quality using the metric defined in \cref{sec:plan-quality}.
These partial-depth sequences extend from the root node and are described as \emph{plan stems} following a similar concept \cite{zimmerman_generating_2002}.
On each iteration of the outer loop, the highest-priority plan stem $\pi$ is removed from the open list (line 6), and a set of extended plans is generated, each with length $|\pi| + 1$, by successively appending each child of the final node of $\pi$ (lines 7--10). The quality of these new plan stems is evaluated, and any that meet the required minimum quality $q$ are added to the open list (lines 11--15).
If $\pi$ was not expanded, then the plan has reached a leaf node, and if it meets the specified diversity requirement $d$, the complete plan is added to the final set of top plans $\mathcal{P}$ (lines 17--19). Ties in plan quality are broken using diversity by continuing to evaluate plans until a lower-quality plan is found (lines 20--28). 
The process is repeated until the open list is empty, or the desired number of plans, $k$, has been generated.

\begin{algorithm}[t!]
    \SetAlgoLined
    \LinesNumbered
    \SetKw{Break}{break}
    \caption{Plan Extraction} \label{algo:plan-extraction}
    \DontPrintSemicolon
    
    \KwIn{$\mathcal{T}$ \AlgComment{root node of tree} \newline
          $Q_{plan}(\pi)$  \AlgComment{quality metric} \newline 
          $D(\pi, \mathcal{P})$  \AlgComment{diversity metric} \newline 
          $k \; \in \; \mathbb{Z}^+$  \AlgComment{cardinality constraint} \newline                     
          $q \; \in \; [0, 1]$  \AlgComment{quality constraint} \newline
          $d \; \in \; [0, 1]$  \AlgComment{diversity constraint} }
    
    \medskip
    \KwOut{$\mathcal{\mathcal{P}}$ \AlgComment{set of top plans}}

    \medskip
    $\mathcal{P} \gets \emptyset$

    $\pi_\mathcal{T} \gets \langle \mathcal{T} \rangle$  \AlgComment{seed plan containing tree root}

    $\mathcal{O} \gets$ PriorityQueue()  \AlgComment{open list}

    $\mathcal{O}$.push($\pi_\mathcal{T}, \; priority=1$)

    \smallskip
    \While{$\mathcal{O} \neq \emptyset$}
    {
        $\pi \gets \mathcal{O}$.pop-max()  \AlgComment{highest-priority plan stem}

        $\sigma \gets \pi$.get-last()  \AlgComment{last node of plan stem}

        $expanded \gets$ False

        \smallskip
        \ForEach{$\sigma' \in \sigma$\textup{.child-nodes()}}
        {
            $\pi' \gets \pi + \sigma'$  \AlgComment{extended plan stem}

            $q_{\pi'} \gets Q_{plan}(\pi')$  \AlgComment{quality of new plan}

            \smallskip
            \If{$q_{\pi'} \geq q$}
            {
                $\mathcal{O}$.push($\pi', \; priority=q_{\pi'}$)

                $expanded \gets$ True
            }
        }

        \smallskip
        \If{$\neg expanded \; \land \; (D(\pi, \mathcal{P}) \geq d)$}
        {            
            \uIf{$|\mathcal{P}| < k$}
            {
                $\mathcal{P}\textup{.push}(\pi)$
                \AlgComment{add to set of top plans}           
            }            
            \Else            
            {
                $q_{min} \gets \textup{min}_{x \in \mathcal{P}}Q_{plan}(x)$

                \uIf{$(d = 0) \lor (Q_{plan}(\pi) < q_{min})$}
                {
                    \Break
                }


                $\mathcal{P}_{qmin} \gets \lbrace x \; | \; x \in \mathcal{P}, Q_{plan}(x) = q_{min}\rbrace$

                $\pi_{dmin} \gets \textup{argmin}_{x \in \mathcal{P}_{qmin}}D(x, \mathcal{P} - x)$
                
                \If{$D(\pi, \mathcal{P}) > D(\pi_{dmin}, \mathcal{P} - \pi_{dmin})$}
                {
                    $\mathcal{P}\textup{.replace}(\pi_{dmin}, \pi)$  \AlgComment{break quality tie}
                }
            }
        }
    }

    \smallskip
    \Return{$\mathcal{P}$}
\end{algorithm}

\begin{lemma}
    At each step of the algorithm, the open list $\mathcal{O}$ contains a set of plan stems guaranteed to produce one or more complete plans with quality $Q_{plan}(\psi \in \mathcal{O}) \geq q$.
\end{lemma}
\begin{proof}
    Let $\sigma$ be the final node of a plan stem $\pi$ in $\mathcal{O}$ with plan quality $Q_{plan}(\pi) \geq q$. Since the $q$-value of node $\sigma$, $Q(\sigma)$, is equal to the weighted average (\cref{eq:node-value-average}) or maximum (\cref{eq:node-value-deterministic}) of the $q$-values of its child nodes, there necessarily exists at least one child $\sigma'$ such that $Q(\sigma') \geq Q(\sigma)$. By \cref{eq:plan-quality}, the quality of the extended plan corresponding to the best child is equal to $Q_{plan}(\pi)$. Applied recursively, a sequence $\psi$ must exist extending to a leaf node with quality $Q_{plan}(\psi) = Q_{plan}(\pi)$, and hence each plan stem in $\mathcal{O}$ is guaranteed to produce at least one complete plan with quality $Q_{plan}(\psi) \geq q$.
\end{proof}

\subsection{Theoretical Analysis}
We describe \cref{algo:plan-extraction} as sound and complete \emph{for a given search tree} if the set of returned plans satisfies the conditions for top-$k$, top-quality or diverse planning as defined in \cref{sec:planning-defs}. That is, the algorithm is sound if each plan in the returned set is guaranteed to meet the requirements for quality and diversity, and complete if every such plan in the tree is guaranteed to be found.

\begin{theorem1}[Soundness] \label{thm:algo-soundness}
    The plan extraction process presented in \cref{algo:plan-extraction} is sound with respect to a pre-generated search tree, such that no plan of higher quality can exist in the tree outside of the returned plan set.
\end{theorem1}
\begin{proof}
    By contradiction: Assume that there exists in the tree some plan $\psi$, with quality $Q_{plan}(\psi)$ that is greater than $Q_{plan}(\pi)$, the quality of a plan $\pi$ in the top plan set. Recall that a plan corresponds to a sequence of nodes starting at the tree root, and hence that $\psi$ and $\pi$ must share a common branching point. 
    From \cref{algo:plan-extraction}, a new plan is added to the open list for each child of the node at the branching point with sufficient corresponding plan quality (lines 9--15). Following \cref{lem:subsequence-quality}, any subsequence of $\psi$, including all possible prefixes, must have quality greater than or equal to $Q_{plan}(\psi)$. Since $Q_{plan}(\psi) > Q_{plan}(\pi)$, a prefix of $\psi$ must exist in the open list with quality greater than the quality of the plan $\pi$.
    The outer loop of \cref{algo:plan-extraction} selects plans from the open list in order of plan quality, and since a prefix of $\psi$ with quality $Q_{plan}(\psi) > Q_{plan}(\pi)$ exists in the list, it is necessarily selected before $\pi$. Thus $\psi$ must be added to the top plan set before $\pi$, so the assumption is not true, and the theorem holds.
\end{proof}

\begin{theorem1}[Completeness]
    The plan extraction process presented in \cref{algo:plan-extraction} is complete with respect to a pre-generated search tree, such that the set of returned plans contains every top plan in the tree that meets the requirements for quality and diversity.
\end{theorem1}
\begin{proof}
    From \cref{lem:subsequence-quality}, for each top plan in the search tree, every subsequence must have greater or equal quality. Since the open list is populated by expanding nodes starting at the root node, a prefix of each top plan is necessarily added to the open list on the first iteration, and by \cref{thm:algo-soundness} the complete plan must eventually be added to the set of top plans before any plan with lower quality. The outer loop continues until either $k$ plans have been generated or the open list is empty, and hence all top plans in the tree are extracted by the algorithm.
\end{proof}

It has been demonstrated that given an infinite number of episodes, the probability of UCB1 selecting a suboptimal action at any given node converges to zero \cite{kocsis_bandit_2006}, and MCTS with UCB1 is hence optimal.
It follows that if \cref{algo:plan-extraction} is sound and complete with respect to a given search tree, and MCTS converges to the optimal policy given sufficient time and memory, then \cref{algo:plan-extraction} is also sound and complete with respect to the complete decision problem.

\subsection{Complexity Analysis} \label{sec:complexity}
The worst-case complexity for top-$k$ plan extraction given a pre-generated search tree is $O(k \cdot d)$, where $d$ is the depth of the tree. Since the algorithm extracts plans in order of quality, the first $k$ iterations of the loop produce the top-$k$ plans, and each iteration expands at most $d$ nodes. The complexity for top-quality is $O(p \cdot d)$, where $p$ is the number of leaf nodes (paths) in the tree and $d$ is the depth. However, this worst case is unlikely in practice, and will occur only if all plans in the tree are above the quality limit. The algorithm assesses the plans in order of quality, so if there are $k$ plans with a quality above the threshold, the algorithm assesses $k \cdot d$ nodes; however, we do not know $k$ a priori. 
For diverse planning, the worst-case complexity is $O(p \cdot d)$ -- at worst, we are required to assess all paths in the tree.
For all of these, in practice, since the tree is fully traversed at least once during its creation, even the worst case represents a fraction of the planning process. We provide empirical results in \cref{sec:experiments}.

\section{Diverse Planning with MCTS}

When applied to top-$k$ or top-quality planning problems, the algorithm presented in the previous section bounds the set of generated solutions according to a cardinality constraint $k$, a quality constraint $q$, or both.
This approach guarantees the generation of a set of plans such that no plan of greater quality exists outside of the returned set.
However, these plans are not guaranteed to differ significantly from one another, and may represent minor variations of the top plan, particularly given MCTS biases exploration toward the most promising branches of the game tree.
In some applications this is desirable, such as plan recognition \cite{sohrabi_state_2017} or plan repair \cite{fox_plan_2006}. In other situations, such as course of action generation or travel planning \cite{myers_generating_1999}, the user desires a set of plans that are \emph{meaningfully} distinct according to some qualitative or quantitative measure, to provide greater coverage of the solution space.

\subsection{Diverse Plan Extraction}
To enforce diversity in the plan set generated by \cref{algo:plan-extraction}, each complete plan is evaluated against the growing set of top plans, and any that do not meet the required diversity are rejected (line 17).
The diversity of a candidate plan, $D(\pi, \mathcal{P})$, is a measure of its dissimilarity to the set of plans $\mathcal{P}$. Plan diversity is calculated from the \emph{minimum pairwise distance} relative to $\mathcal{P}$, which may be based on qualitative measures that are specific to the domain, or quantitative measures that are domain-independent.

A number of domain-independent plan distance measures have been presented in previous works, typically comparing states, actions, causal links, or landmarks \cite{srivastava_domain_2007,bryce_landmark-based_2014}, requiring varying degrees of problem and domain theory.
Given a basis for comparison, each plan is aggregated as either an unordered set or a position-sensitive sequence, and a difference is calculated between each pair of plans.
We assume a distance measure is provided by the user, and for the remainder of this paper employ a simple diversity metric based on state set distance:
    \begin{equation} \label{eq:state-difference}
        \delta(\pi, \psi) = \frac{|\pi - \psi|}{|\pi|}
    \end{equation}
where $\pi$ and $\psi$ are plans, and $\pi - \psi$ is a one-way set difference. That is, $\delta(\pi, \psi)$ is the fraction of unique states in $\pi$ that are not in $\psi$ (for simplicity, a plan here represents a set of states).
The diversity metric used when considering a new plan $\pi$ in \cref{algo:plan-extraction} is then its minimum distance from any plan in the existing plan set $\mathcal{P}$:
    \begin{equation} \label{eq:state-diversity}
        D(\pi, \mathcal{P}) = \min_{\psi \in \mathcal{P}} \delta(\pi, \psi)
    \end{equation}

\begin{theorem1}
    The diverse plan set generated by \cref{algo:plan-extraction} is Pareto-optimal: there exists no alternative set that dominates the returned set with respect to quality and diversity.
\end{theorem1}
\begin{proof}
    From \cref{algo:plan-extraction}, plans are added to the final plan set in order of quality (lines 6 and 19), breaking ties using diversity (lines 20--28), and bounded by the diversity constraint $d$ (line 17). It follows that every plan $\pi \in \mathcal{P}$ has quality no less than any plan $\psi \notin \mathcal{P}$ with diversity $D(\psi, \mathcal{P}) \geq d$, and hence there is no set outside of $\mathcal{P}$ that can increase quality without decreasing diversity, or vice versa. Therefore, the set is Pareto-optimal with respect to quality and diversity.
\end{proof}

\subsection{Diverse Multi-Armed Bandit Policies}
The diverse-planning approach described in the previous section collects diverse plans from a pre-generated search tree, and as a result relies on sufficient diversity in the planning process used to build the tree. MCTS naturally biases exploration toward the most promising areas of the search space. As a result, the highest-quality branches in the tree are likely to be clustered with the optimal plan. While this side effect may be desirable for top-$k$ and top-quality planning, it may lead to starvation of branches that are distant from the optimal plan, and hence reduce the diversity of the final plan set.

To address this, we propose two approaches to increasing the diversity of the pre-generated MCTS tree.
The first and most obvious solution is to increase the bias constant that balances exploration and exploitation in the selection policy used during construction of the tree (e.g. $C$ in \cref{eq:ucb1}), to encourage exploration of less-promising areas of the solution space. This approach incurs a performance penalty, sacrificing tree depth for breadth and hence requiring a higher number of iterations for the MCTS tree to produce reliable value estimates. Furthermore, the additional exploration is undirected and does not necessarily bias the search toward high-quality diverse plans.

To direct tree exploration toward diverse solutions, we propose a second approach that extends the bandit policy by explicitly considering diversity. During the selection phase of the MCTS algorithm, the value of each node $\sigma$ considered by the multi-armed bandit policy is augmented with the diversity of the corresponding plan stem $\pi_\sigma = \{\sigma_0, \dots, \sigma\}$.
For example, the UCB1 formula (\cref{eq:ucb1}) may be modified to include an additional term:
    \begin{equation}
        \text{DiverseUCB1}(\sigma) = Q(\sigma) + C \sqrt{\frac{2 \ln n}{n_\sigma}} + D(\pi_\sigma, \mathcal{P})
    \end{equation}
where $D(\pi_\sigma, \mathcal{P})$ is the diversity of the plan stem $\pi$ relative to a set of plans $\mathcal{P}$.
Similar to \cref{algo:plan-extraction}, plan diversity is defined relative to a bounded set of high-quality plans extracted from the partially-constructed tree, in this case updated dynamically as the tree is constructed.
Exploration of this approach is left for future work.

\section{Experimental Validation} \label{sec:experiments}

To evaluate our approach, we test the plan extraction algorithm over a path-planning problem with hidden information.
This experiment is provided as a proof of concept for generating plan \emph{sets} over black-box simulators, in the absence of a benchmark for such problems.
In the problem scenario, a mission planner must deliver medical supplies on a battlefield using a set of delivery drones, while avoiding hostile forces.
The planner is provided with a map and goal location, but the locations of enemy combatants are unknown. Enemy combatants will shoot down the drones if they pass nearby.
The scenario objective is to generate a set of plans that maximises the probability of at least one drone successfully reaching the target.
The source code for the simulator and experiments is available at \href{http://bit.ly/3WnQxme}{http://bit.ly/3WnQxme}.

Three configurations of the algorithm are considered: a top-$k$ planner, a top-quality planner, and a diverse planner. Each of the planners generates a set of plans for up to five drones (i.e. $k \leq 5$). The top-quality planner restricts the plans to a minimum 80\% of the quality of the optimal plan ($q = 0.8$), while the diverse planner enforces a minimum 50\% plan distance ($d = 0.5$). For the purposes of these experiments, a relatively simple diversity metric is used, based on state set difference (\cref{eq:state-diversity}).
Two baselines are implemented for comparison: a planner that extracts the single optimum plan for each problem instance, and a random planner that extracts a set of paths selected uniformly from the search tree.

\begin{figure}[!ht]
    \centering
    \begin{subfigure}{0.49\linewidth}
        \includegraphics[width=\linewidth]{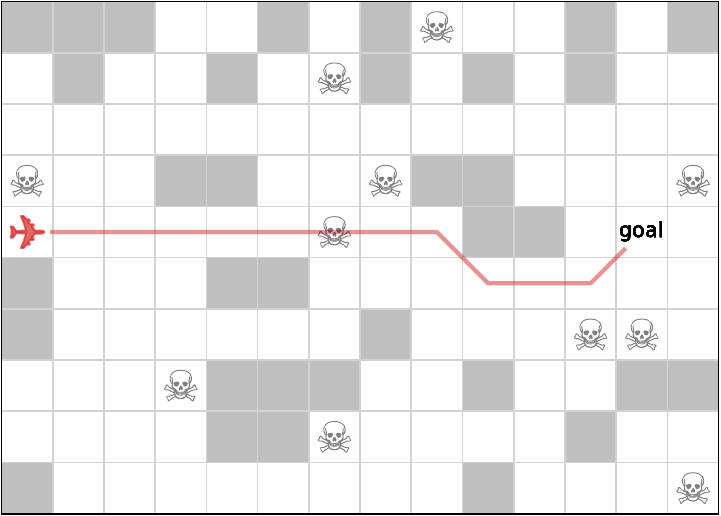}
        \caption{Single planner}
    \end{subfigure}
    \begin{subfigure}{0.49\linewidth}
        \includegraphics[width=\linewidth]{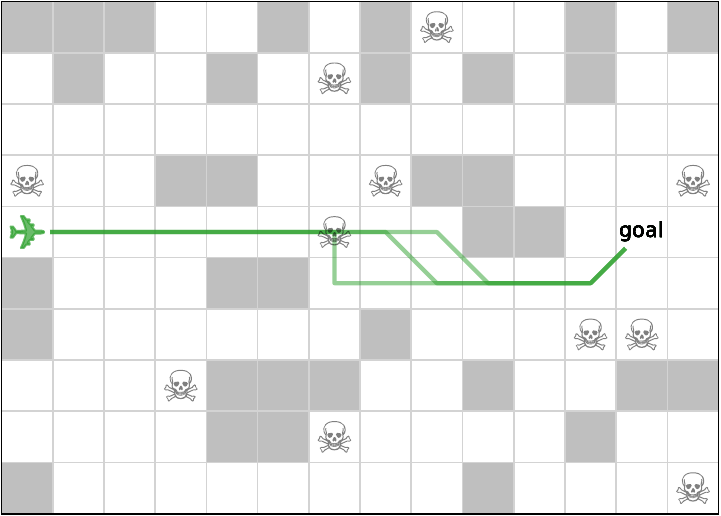}
        \caption{Top-quality planner}
    \end{subfigure}

    \medskip

    \begin{subfigure}{0.49\linewidth}
        \includegraphics[width=\linewidth]{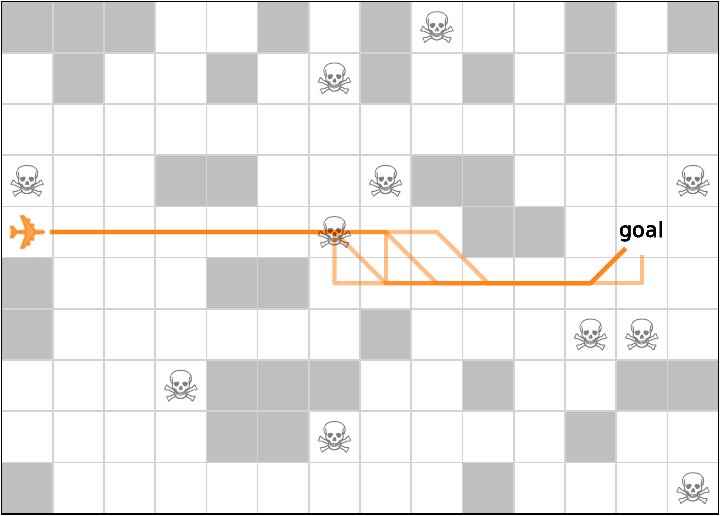}
        \caption{Top-$k$ planner}
    \end{subfigure}
    \begin{subfigure}{0.49\linewidth}
        \includegraphics[width=\linewidth]{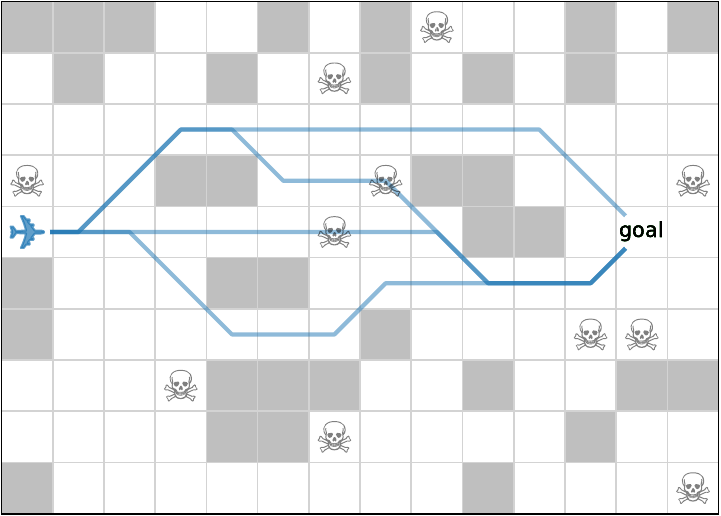}
        \caption{Diverse planner}
    \end{subfigure}    
    \caption{Example of a solution set generated by each planner type. In this example, with risk level 10\%, only the diverse planner is successful.}
    \label{fig:gridworld-paths}
\end{figure}

\begin{figure}[!ht]
    \centering
    \includegraphics[width=1.0\linewidth]{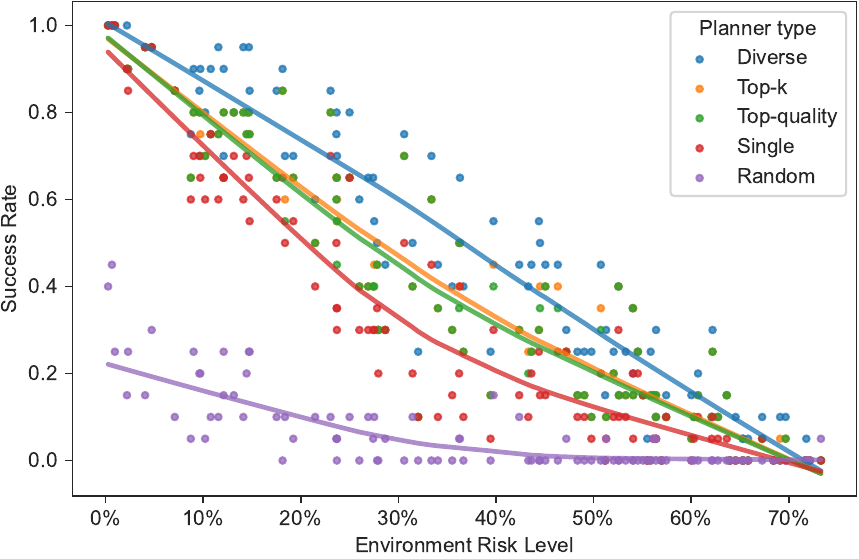}
    \caption{Experimental results for each planner type. The environment risk level is controlled by varying the percentage of map locations occupied by enemy combatants.}
    \label{fig:risk-vs-result-lmplot}
\end{figure}

For the experiment, 2000 randomised problem instances are generated across 100 risk levels. The risk level for each instance is controlled by varying the level of hidden information, represented by the percentage of map locations occupied by enemy combatants.
For each unique problem instance, an MCTS tree is constructed over 20,000 iterations ($\approx 1.3$ seconds on an Intel Core i7-8650U processor), using a simulation model with hostile forces removed to reflect the limited information available to the planners. The plan set generated by each planner type is then executed in the true environment (\cref{fig:gridworld-paths}) where the location of enemy combatants is known. The set of plans is considered  successful if at least one plan in the set reaches the objective without encountering hostile forces. The average success rate for each planner is recorded over 20 replications at each risk level.

\begin{figure}[!ht]
    \centering
    \includegraphics[width=1.0\linewidth]{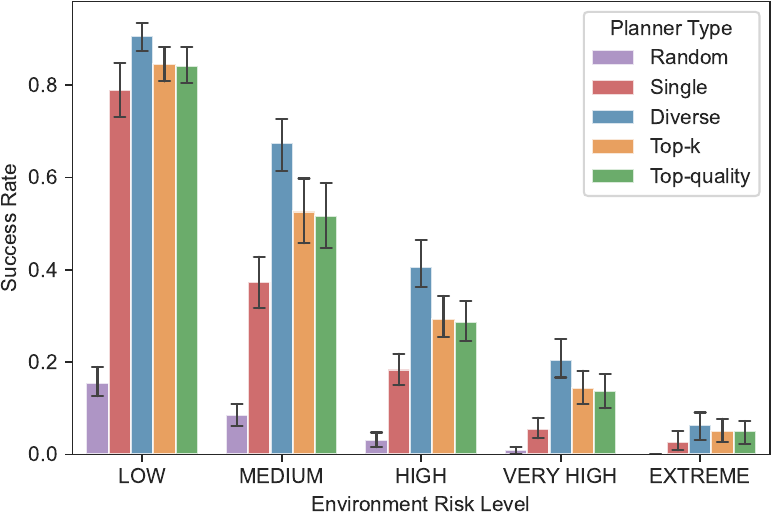}
    \caption{Comparison of mean success rate for each planner (error bars denote 95\% confidence interval).}
    \label{fig:risk-vs-result-barplot}
\end{figure}

\cref{fig:risk-vs-result-lmplot,fig:risk-vs-result-barplot} capture the results. For this domain, our algorithm shows a significant improvement in success rate over each of the baseline approaches. As expected, the single planner baseline performs well when the risk level is low and the optimal path is unlikely to encounter hidden enemy combatants. As the risk level increases, performance for the single planner rapidly decreases relative to the top-$k$, top-quality and diverse planners. At medium to high risk levels, where up to 80\% of map locations are occupied by hostile forces, the diverse planner has a success rate 1.8--3.7$\times$ that of the single planner (95\% CI).
While the difference in success rate between the diverse planner and all other planners is statistically significant, the difference between the top-$k$ and top-quality planners is not. This is to be expected, as the plan sets generated by these planners differ only when there are fewer than $k$ plans that meet the quality requirement $q$.
The poor performance of the random planner baseline suggests that the benefit provided by the top-$k$, top-quality and diverse planners is due to effective plan discovery, rather than simply due to greater coverage of the state space compared to the single planner.

\begin{figure}[!t]
    \centering
    \includegraphics[width=1.0\linewidth]{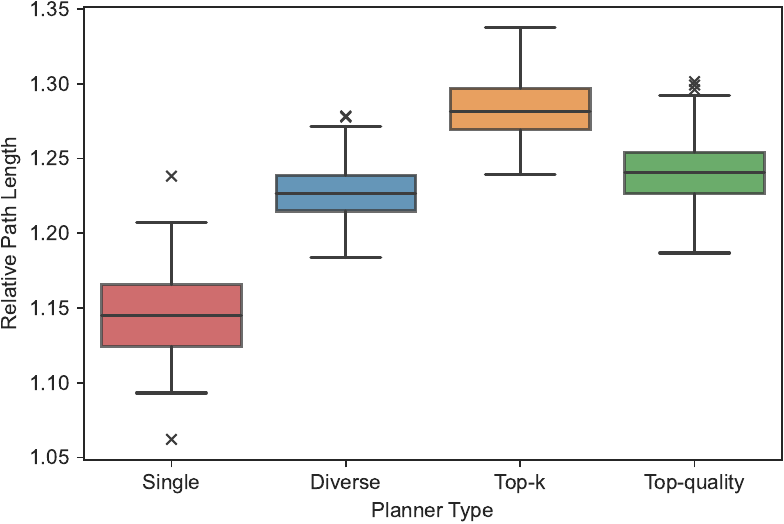}
    \caption{Average path length generated by each planner, relative to the shortest unobstructed path to the objective.}
    \label{fig:pathcost_by_planner_boxplot}
\end{figure}

\cref{fig:pathcost_by_planner_boxplot} demonstrates the average path cost for each planner relative to the shortest unobstructed path between the initial location and the objective. Our algorithm generates plans in order of expected return, and does not explicitly minimise path cost. However, it can be seen that the plans generated by the diverse, top-quality and top-$k$ planners add minimal additional path length over the optimal plan in this application.

Although not proposed as an optimised algorithm, we additionally consider the computational overhead of plan set extraction using our approach. Given an MCTS tree constructed over $1.5s \pm 820ms$ (95\% CI), the time required to generate a plan set during an experiment run ranges from $312\mu s \pm 28.6\mu s$ (95\% CI) for the single planner, to $1.76ms \pm 152.4\mu s$ (95\% CI) for the diverse planner. Compared to the cost of generating the MCTS tree, these costs are considered to be negligible.

\section{Discussion} \label{sec:discussion}

Our experimental results demonstrate that plan extraction from pre-generated search trees is a viable approach to the generation of bounded plan sets.
However, empirical observation suggests that further work is needed to improve the diversity of search trees generated using MCTS algorithms.
Additionally, while a state-set diversity metric was effective for the discrete path-planning domain used for our experiments, previous work has shown that more sophisticated measures are needed for general planning problems \cite{bryce_landmark-based_2014}. In particular, it has been observed that plans that are distant with respect to state and action sequences may be fundamentally the same when considered semantically. A possible solution may lie in sequence comparison methods used in genomics, repurposed to identify causal patterns in agent trajectories \cite{vadakattu_strategy_2023}.

As discussed in \cref{sec:plan-quality}, our definition of plan quality is based on expected return rather than plan cost. As a result, while the plans generated by our approach are guaranteed to have the highest expected return, they are not guaranteed to be the most efficient with respect to other measures such as path length, unless these are integrated into the reward function.
Previous work has explored the trade-off between efficiency and diversity \cite{roberts_evaluating_2014}. Our experiments suggest that where efficiency is an important consideration, the optimal level of plan diversity may be driven by the expected risk level of the environment.
If the risk is low, such as in fully-observable and deterministic domains, then single planning may be optimal. If the risk is intermediate, and the successful solution is expected to be close to the optimal plan, then top-$k$ or top-quality planning may be sufficient. If the risk level is high or unknown, then the additional cost of diverse planning may be justified by an increased probability of success.

As a post-processing method, our algorithm provides a novel capability: regeneration of alternative plan sets without replanning. Previous approaches have generated multiple plans by repeatedly solving modified versions of the planning problem, often incurring significant additional cost over single planning \cite{nguyen_generating_2012,riabov_new_2014,katz_novel_2018,speck_symbolic_2020}. 
In contrast, our approach operates over a pre-generated search tree, and does not require replanning to produce multiple plans.
Once the tree has been generated, new sets of plans can be extracted with different quality and diversity constraints for negligible additional cost.
This feature is particularly useful for planning problems requiring long processing times, as it enables the user to experiment with constraints and diversity measures without incurring the cost of solving the planning problem anew.

\section{Conclusion}

In this paper, we have introduced a novel approach to top-$k$, top-quality and diverse planning over black-box simulators that removes the requirement for a symbolic model of the problem. We have presented a metric for the relative qualities of paths in a search tree, and an algorithm for extracting sets of plans in best-first order, along with a theoretical analysis showing that our approach finds exactly the set of top plans. Further, we have discussed strategies to improve the diversity of generated plans by modifying the multi-armed bandit selection policy used during search. Finally, we have provided an empirical evaluation to demonstrate the practicality of our approach.

There are a number of areas of continuing research, including extending our algorithm to multi-agent problems, and further developing our proposed diverse bandit policy.
In future work we will explore more sophisticated diversity measures, such as causal pattern extraction, to improve the generation of meaningfully diverse sets of plans.

\newpage
\bibliography{references}

\end{document}